\pdfoutput=1
\documentclass[conference]{IEEEtran}
\IEEEoverridecommandlockouts
\usepackage{epstopdf}
\usepackage{epsfig}
\usepackage{booktabs}
\usepackage{graphicx}
\usepackage{multirow}
\usepackage{amsmath,amssymb,amsfonts}
\usepackage[lined,boxed,vlined,ruled,linesnumbered]{algorithm2e}
\usepackage{caption}
\usepackage{tikz}
\usepackage{pgflibraryarrows}
\usepackage{pgflibrarysnakes}
\usepackage{subfigure}
\usepackage[utf8]{inputenc}
\usepackage[english]{babel}
\usepackage{hyperref}
\usepackage{dsfont}
\usepackage{bbm}
\usepackage{url}
\usepackage{makecell}
\usepackage{mathrsfs}
\usepackage{balance}
\usepackage{xcolor}
\usepackage{cite}
\usepackage{algorithmic}
\usepackage{textcomp}

\def\BibTeX{{\rm B\kern-.05em{\sc i\kern-.025em b}\kern-.08em
    T\kern-.1667em\lower.7ex\hbox{E}\kern-.125emX}}

\newcommand{\stitle}[1]{\vspace{1ex}\noindent\textup{\textbf{#1}}}

\newtheorem{definition}{Definition}
\newtheorem{proof}{Proof}

\begin{document}

\title{An End-to-End Deep RL Framework for Task Arrangement in Crowdsourcing Platforms}

\author{
{Caihua Shan{$\,^\dag$}~~~Nikos Mamoulis{$\,^\S$}~~~Reynold Cheng{$\,^\dag$}~~~Guoliang Li{$\,^\#$}~~~Xiang Li{$\,^\dag$}~~~Yuqiu Qian{$\,^\ddag$}}

\vspace{1.6mm}\\
{$\,^\dag$}\, The University of Hong Kong, 
{$\,^\S$}\, University of Ioannina,
{$\,^\#$}\,  Tsinghua University,
{$\,^\ddag$}\, Tencent Inc.
\vspace{1.6mm}\\
\{chshan, ckcheng, xli2\}@cs.hku.hk, ~nikos@cs.uoi.gr, ~liguoliang@tsinghua.edu.cn, ~yuqiuqian@tencent.com

}

\maketitle

\begin{abstract}

In this paper, we propose a Deep Reinforcement Learning (RL) framework for task arrangement, which is a critical problem for the success of crowdsourcing platforms. Previous works conduct the personalized recommendation of tasks to workers via supervised learning methods. 
However, the majority of them only consider the benefit of either workers or requesters independently. In addition, they cannot handle the dynamic environment and may produce sub-optimal results. To address these issues, we utilize Deep Q-Network (DQN), an RL-based method combined with a neural network to estimate the expected long-term return of recommending a task. DQN inherently considers the immediate and future reward simultaneously and can be updated in real-time to deal with evolving data and dynamic changes. Furthermore, we design two DQNs that capture the benefit of both workers and requesters and maximize the profit of the platform. To learn value functions in DQN effectively, we also propose novel state representations, carefully design the computation of Q values, and predict transition probabilities and future states. Experiments on synthetic and real datasets demonstrate the superior performance of our framework.

\end{abstract}

\begin{IEEEkeywords}
crowdsourcing platform, task arrangement, reinforcement learning, deep Q-Network
\end{IEEEkeywords}

\section{Introduction}\label{sec:introduction}

Crowdsourcing is an effective way to address computer-hard tasks by utilizing numerous ordinary human (called {\em workers} or {\em the crowd}). In  commercial crowdsourcing platforms (i.e., Amazon MTurk~\cite{AMT} or CrowdSpring~\cite{CrowdSpring}), 
requesters first publish tasks with requirements (e.g., collect
labels for an image) and awards (e.g., pay $0.01$ per labeling). When a worker arrives, the platform shows him/her a list of available tasks (posted by possibly different requesters), which are ordered by a certain criterion, e.g., award value or creation time. The worker can select any of the tasks in the list based on summary information for each task, such as the title, the description and the award. Finally, s/he clicks on a task, views more detailed information and decides whether to complete it or not.

\begin{figure}[t!]
 \centering  
 \includegraphics[width=0.45\textwidth]{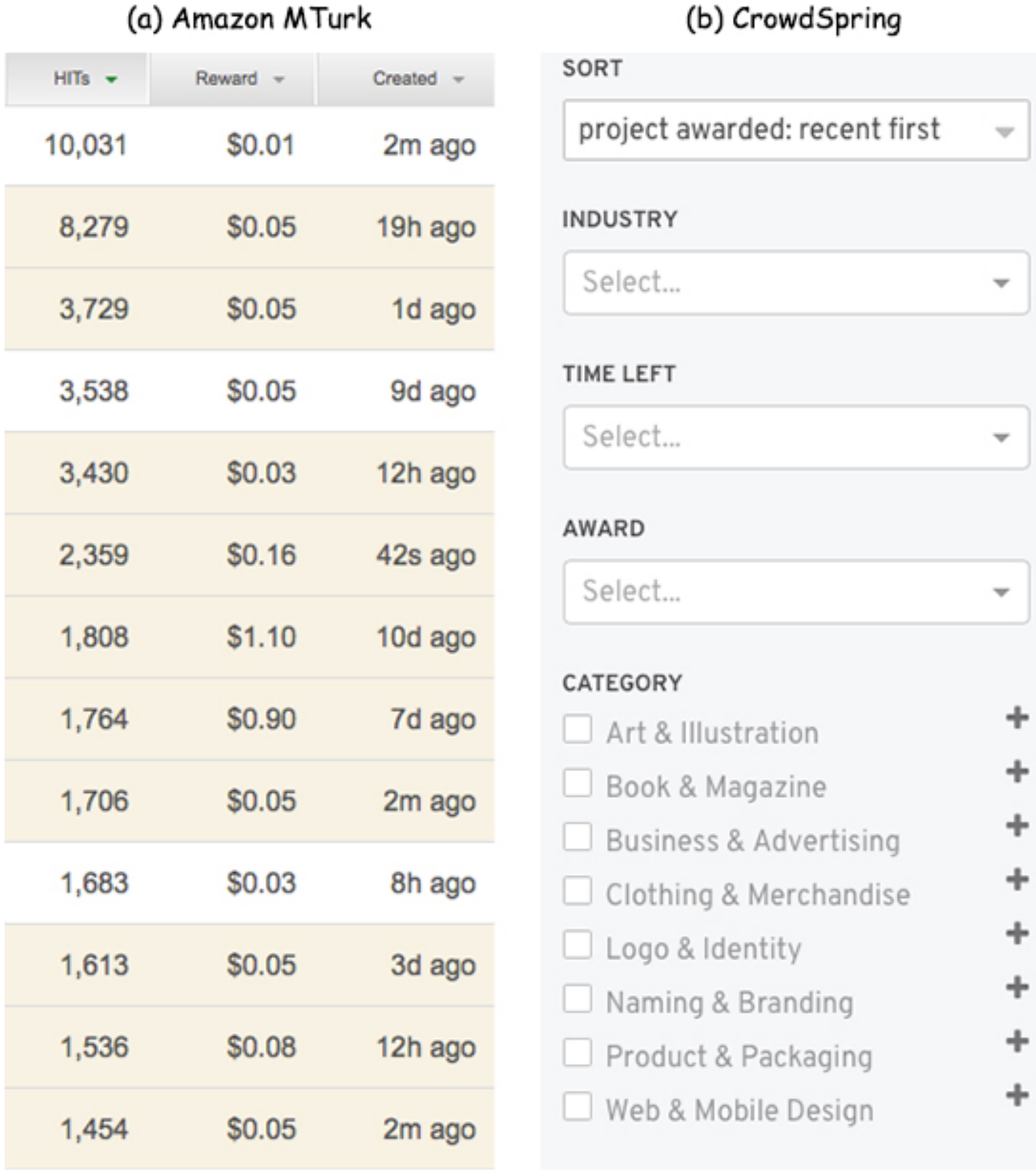}
 \caption{Sorting or Filtering Functions}  
\label{fig:HIT_list} 
\end{figure}

As shown in Fig.~\ref{fig:HIT_list}, the current platforms only provide a simple sorting or filtering function for tasks, i.e., sorting by creation time, filtering by category, etc. Due to the large number of available tasks, previous work\cite{safran2018efficient,taskrec} pointed out that manually selecting a preferred task is tedious and could weaken workers' enthusiasm in crowdsourcing. 
They propose some supervised learning methods (e.g., $k$NN classification or probabilistic matrix factorization) to conduct personalized recommendation of tasks to workers. However, these approaches come with several shortcomings.

First of all, previous works only consider the recommendation and assignment
of tasks having as target the 
individual benefit of 
either the workers or the requesters. 
If we only consider the workers' preferences or skills, some tasks in domains of rare interest cannot find enough workers to complete.
On the other hand, if we only consider the benefit of the requesters, i.e., collecting high-quality results by a given deadline, the assignment of tasks might be {\em unfair} to workers, lowering their motivation to participate.
The goal of a commercial platforms is to maximize the number of {\em completed} tasks, as they make a profit by receiving a commission for each such task.
To achieve this, they should attract as many tasks as possible by requesters and as many as possible workers to complete these tasks. 
Hence, it is necessary to {\em balance} the benefit of both of workers and requesters by satisfying the objectives of both parties
to the highest possible degree. 

Second, previous works are not designed for 
handling real dynamic environments. 
New tasks are created and old tasks expire all the time. The {\em quality} of a given task (e.g., accuracy of labeling) also keeps changing as it gets completed by workers. Besides, we do not know which worker will come at the next moment, and the workers’ preferences are evolving based on the currently available tasks. The models based on supervised learning cannot update the preferences of workers in real-time. 
We show by experimentation that, even if we update supervised learning-based models every day, their performance is still not satisfactory. 

Further, the majority of existing works are designed for maximizing the immediate (short-term) {\em reward}, i.e., select the task with the maximum predicted completion rate for the coming worker, or choose the task that yields the maximum quality gain. They disregard whether the recommended tasks will lead to the most profitable (long-term) reward in the future; hence, they may generate suboptimal suggestions w.r.t. the long-term goal. 

To address the above issues, we propose a {\em Deep Reinforcement Learning} framework for task arrangement in this paper.  We model the interactions between the environment (workers and requesters) and the agent (the platform) as a Markov Decision Process (MDP).
We apply {\em Deep Q-Network} (DQN), a widely used reinforcement learning method, training a neural network to estimate the reward for recommending each task.
DQN naturally considers the immediate and future reward simultaneously in the online environment (i.e., the continuing coming workers and changing available tasks). 
Besides, DQN can be updated in real-time after each worker's feedback, seamlessly handling dynamic and evolving workers and tasks.

Despite the advantages of DQN  in crowdsourcing platforms, 
it cannot be directly applied into our task arrangement problem. 
A typical DQN for recommendation systems only models the relationship between users and items, i.e., workers and tasks in our context.
Here,
we should also take into consideration
the relationships
among all available tasks.
To capture all the information of the environment, we design a novel state representation that concatenates the features of workers and currently available tasks, as well as a particular Q-Network to handle the set of available tasks with uncertain size and permutation-invariant characteristics. 

Besides, workers and requesters have different benefits, and we choose to use two MDPs to model them. 
If we only consider to recommend
tasks of interest for workers,
the actions decided by the MDP for a worker are independent to those for other workers.
However, 
the assigned tasks and the corresponding feedback of previous workers do affect the action assigned to the next worker
and the quality of tasks (i.e.,
the benefit of requesters).
Thus, we design two separate DQNs to represent these two benefits and then combine them.

Furthermore, DQN is a model-free method which computes the transition probability of (future) states implicitly. Since such (future) state is composed of the (next) coming workers and the available tasks, 
these workers and tasks could
generate a large number of state representations and
thus very
sparse transitions between states.
This further
leads to possibly inaccurate estimation of transition probability and slow convergence.
To address such problem,
we revise the equation of computing Q values,
and predict transition probabilities and future states explicitly,
after obtaining the feedback from a worker.  
Specifically, we utilize the worker arrival distribution (which will be discussed in Sec.~\ref{sec:future_state_w} and Sec.~\ref{sec:future_state_r}) 
to predict the probability when the next timestamp is, who the next worker is, and how many tasks are available.


Our contributions can be summarized as follows:

1) To the best of our knowledge, we are the first to propose a Deep Reinforcement Learning framework for  task arrangement in crowdsourcing platforms. 

2) We apply a Deep Q-Network (DQN) to handle both immediate and future rewards, aiming at optimizing a {\em holistic objective} from the perspectives of both workers and requesters in the long term.

3) We design a novel and efficient state representation, revise equations for computing Q values and predict transition  probabilities and future states explicitly.

4) We use both synthetic and the real datasets to demonstrate the effectiveness and efficiency of our framework. 

The rest of the paper is organized as follows. We define the problem, formulate the MDPs and introduce Deep Q-Network in Sec.~\ref{sec:model}. In Sec.~\ref{sec:overview}, we describe the entire process of our framework. Its modules are described in detail in Sec.~\ref{sec:method_ben_workers}, \ref{sec:method_ben_requesters} and \ref{sec:integration}. Experiments on synthetic and real data  are conducted in Sec.~\ref{sec:experiments}. We discuss related work in Sec.~\ref{sec:relatedwork} and conclude in Sec.~\ref{sec:conclusions}.

\section{Problem Statement}\label{sec:model}

\begin{figure*}[t!]
  \centering
   \includegraphics[width=0.7\textwidth]{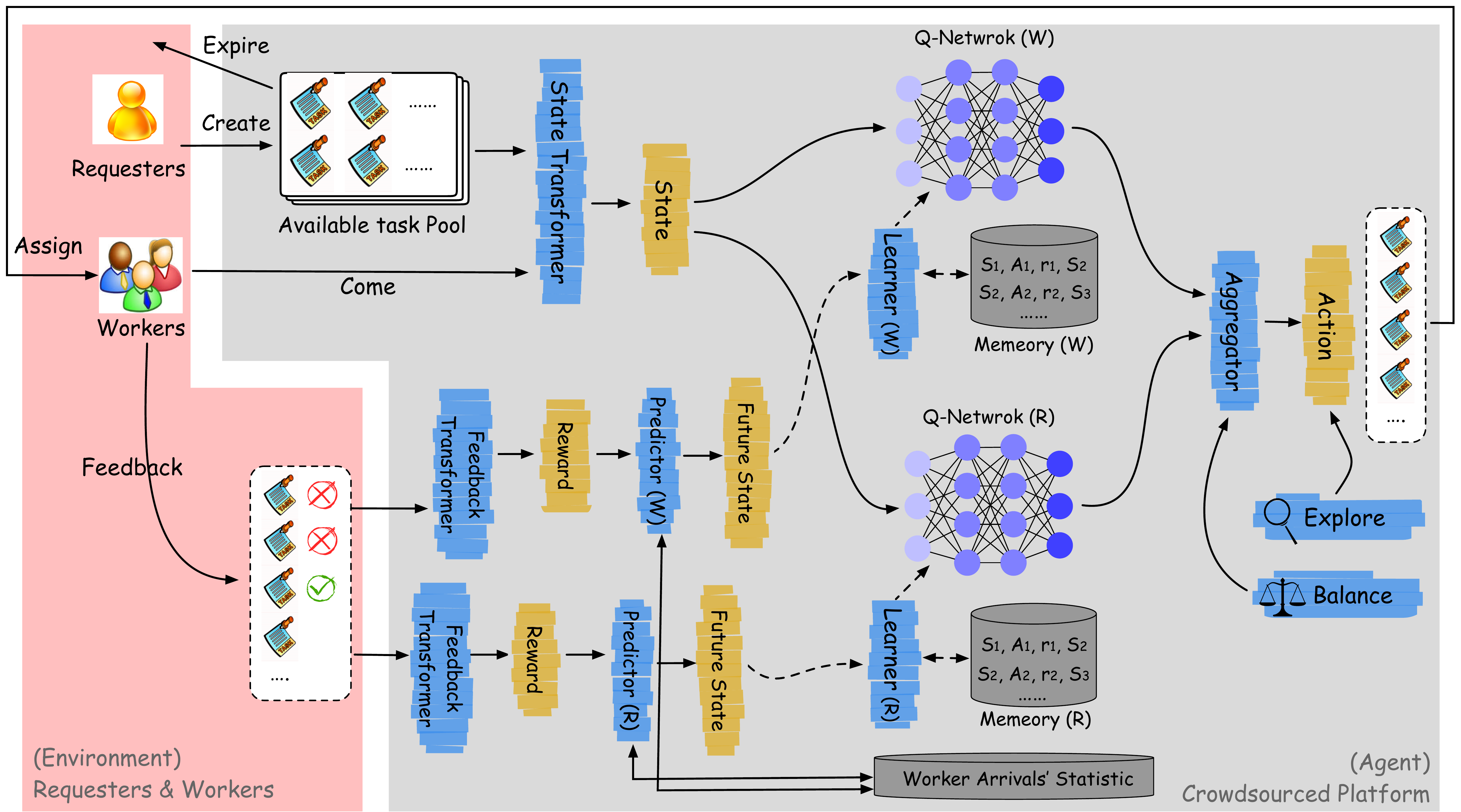}
  \caption{DRL Framework}
  \label{fig:framework} 
\end{figure*}

\subsection{Problem Definition}


The goal of the proposed task arrangement system is to \textbf{assign a task or recommend a sorted list of tasks to a coming worker},
which benefits both workers and requesters.
The system should cope with dynamic changes and is required to interact in real-time.

\subsection{Problem Formulation as MDPs} 

Here we model the task arrangement problem as a reinforcement learning problem, by defining two MDPs.
While the crowdsourcing platform (the {\it agent}) interacts with requesters and workers (the {\it environment}),
requesters influence the pool of available tasks  by setting the start date and a deadline of tasks and obtaining the result of each task after its deadline.
The agent does not need to take any action.
Thus,
we mainly consider the actions by workers. 

\stitle{MDP(w) (for the benefit of workers):}
Following the MDP setup of a typical item recommendation system \cite{zheng2018drn, zhao2018recommendations}, our MDP considers the benefit of workers as follows.
At each timestamp $\{1,2,\cdots,i\}$, a worker $w_i$ comes and there is a set of available tasks $\{T_i\}$ posted by requesters.
\begin{itemize}
 \item State $s_i$ is defined as the recent completion history of $w_i$, i.e., the representation of the state is the {\em feature} of the worker $w_i$, $f_{s_i} = f_{w_i}$.
 \item An action $a_i$ is to recommend some of the available tasks to $w_i$.
 There are two kinds of actions based on the problem setup. If the problem is to recommend one task, the possible actions are all available tasks, i.e., $a_i = t_j, \forall t_j \in \{T_i\}$. If the problem is to recommend a sorted list of tasks, possible actions are all possible permutations of available tasks, where $a_i = \sigma(T_i) = \{t_{j_1},t_{j_2}...\}$ and $\sigma$ is a rank function.
 \item Reward $r_i$ is decided by the feedback of $w_i$ given $(s_i, a_i)$. 
 $r_i=1$ if $w_i$ completes a task. Otherwise $r_i = 0$.   
 \item Future State $s_{i+1}$ happens when the same worker $w_i$ comes again. The worker feature $f_{w_i}$ is changed if $r_i>0$. Thus $f_{s_{i+1}}$ is the updated worker feature $f_{w_i}$ by $r_i$, i.e., the feature of worker $w_i$ when $w_i$ comes again.
 \item Transition $Pr(s_{i+1}|s_i,a_i,r_i)$ is the probability of state transition from $s_{i}$ to $s_{i+1}$, which depends on the success ($r_i$) of completing a certain task of $a_i$ by $w_i$.
 \item The discount factor $\gamma \in [0,1]$ determines the importance of future rewards compared to the immediate reward in reinforcement learning.
\end{itemize}

Based on the MDP(w) definition, the global objective is to maximize the cumulative completion rate of workers in the long run.

\stitle{MDP(r) (for the benefit of requesters):}
Again, each timestamp $i$ is triggered by the coming worker $w_i$ and there exists a set of available tasks $\{T_i\}$. However, as we now consider the sum of the qualities of tasks posted by requesters, some elements of the MDP are different:
\begin{itemize}
 \item State $s_i$ is defined as the previous completion history of $w_i$ and currently available tasks $\{T_i\}$. The worker quality $q_{w_i}$ and the task quality $q_{t_j}, \forall t_j \in \{T_i\}$ are also considered. $f_{s_{i}}$ is the combination of all these features, i.e.,  $f_{s_{i}}=[f_{w_i},f_{T_i},q_{w_i},q_{T_i}]$. 
 \item Action $a_i$ is the same as in MDP(w). 
 \item Reward $r_i$ is decided by the feedback of $w_i$ given $(s_i, a_i)$. $r_i$ is the quality gain of the completed task by $w_i$. If $w_i$ skips all the recommended tasks, $r_i = 0$. 
 \item Future State $s_{i+1}$ happens when the next worker $w_{i+1}$ comes, no matter whether $w_{i+1} \neq w_i$. The worker feature $f_{w_i}$ and the quality of completed task $q_{t_{j'}}$ may be changed if $r_i>0$. 
 \item Transition $Pr(s_{i+1}|s_i,a_i,r_i)$ depends on the success and quality gain ($r_i$) of completing a certain task of $a_i$ by $w_i$. Moreover, it is related to the next worker $w_{i+1}$. 
 \item The discount factor $\gamma$ is the same as in MDP(w). 
\end{itemize}
According to the MDP(r) definition, the global objective is to maximize the cumulative quality gains of tasks in the long run.

\stitle{Remark:} 
The reason why we use different definitions of states is that we have different global objectives. To optimize the workers' benefits, we are supposed to explore and exploit the relationship between each worker and each task. Through trial-and-error recommendations, we can automatically learn the optimal strategy for each worker, even if the interest of workers is evolving. 
However, maximizing the sum of the quality of tasks is similar to solving a matching problem. We not only need to consider the worker-task relationships, but also all available tasks to obtain the overall maximum benefit. So a state in MDP(r) is composed by the worker and the currently available tasks. 

To unify the state definition in two MDPs, we use the state definition of MDP(r) in place of MDP(w) since they have an inclusion relation. Thus the state in MDP(w) is also composed by $w_i$ and $T_i$ and its representation becomes $f_{s_i} = [f_{w_i},f_{T_i}]$.

\subsection{RL and Deep Q-Network} \label{sec:RL_basic}
\subsubsection{Q-Learning}
Q-learning \cite{Q_learning} is a value-based and model-free reinforcement learning algorithm, which defines two value functions to find the optimal policy $\pi: \mathcal{S} \to \mathcal{A}$ that maximizes the cumulative reward. 
$V^{\pi}(s)$ is the state value function where 
$V^{\pi}(s) = \mathbb{E}[\sum_{i=0}^{\inf} \gamma^{i} r_i | s_0 = s, \pi ]$
is the expected return following the policy $\pi$ given the state $s$.
Similarly, the state-action value function $Q^{\pi}(s,a)$ is the expected return given state $s$ and action $a$, where
$Q^{\pi}(s,a) = \mathbb{E}[\sum_{i=0}^{\inf} \gamma^{i} r_i | s_0 = s, a_0=a, \pi ].$

Based on  Bellman's equation \cite{sutton2018reinforcement}, the optimal Q value function $Q^{*}(s,a)$ with the optimal policy satisfying
$$
Q^{*}(s_i,a_i) = \mathbb{E}_{s_{i+1}}[r_i+ \gamma \max_{a'} Q^{*}(s_{i+1},a')|s_i,a_i].
$$
Thus, it learns $Q(s_i,a_i)$ iteratively by choosing the action $a_i$ with the maximum $Q(s_i,a_i)$ at each state $s_i$. Then it updates $Q(s_i,a_i) \leftarrow (1-\alpha)Q(s_i,a_i) + \alpha (r_i+\gamma \max_{a'} Q(s_{i+1},a'))$ where $\alpha \in [0,1]$ is the learning rate.

\subsubsection{Deep Q-Network}
In practice, we may have enormous state and action spaces, making it impossible to estimate $Q^{*}(s,a)$ for each $s$ and $a$. Besides, it is hard to store and update so many state-action pairs. It is typical to use a highly nonlinear and complex function to approximate, i.e., $Q^{*}(s,a) \approx Q(s,a;\theta)$.
Hence, {\em Deep Q-Network}\cite{double_q_learning} is proposed, which uses a neural network with parameters $\theta$ as the Q-network. It is learned by minimizing the mean-squared loss function as follows:
\begin{equation} 
\begin{aligned}
L(\theta) & = \mathbb{E}_{\{(s_i,a_i,r_i,s_{i+1})\}}[(y_i-Q(s_i,a_i;\theta))^{2}] \\
y_i & =  r_i+\gamma \max_{a_{i+1}} Q(s_{i+1},a_{i+1};\theta)
\end{aligned}
\end{equation}
where $\{(s_i,a_i,r_i,s_{i+1})\}$ is the historical data, stored in a large memory buffer sorted by occurrence time. By differentiating the loss function with respect to $\theta$, the gradient update can be written as:
\begin{equation}
\begin{aligned}
\nabla_{\theta} L(\theta) = &  \mathbb{E}_{\{(s_i,a_i,r_i,s_{i+1})\}}[(r_i+ \gamma \max_{a_{i+1}} Q(s_{i+1},a_{i+1};\theta) \\
     & - Q(s_i,a_i|\theta)) \nabla_{\theta}Q(s_i,a_i|\theta)]
\end{aligned}
\end{equation}
In practice, 
{\em stochastic gradient descent}
can be used to efficiently optimize the loss function.


\section{Overview} \label{sec:overview}
Fig.~\ref{fig:framework} illustrates the whole framework. A worker $w_i$ comes and sees a set of available tasks $\{T_i\}$ posted by requesters at timestamp $i$. The representation of a state includes the feature of worker $w_i$ and the available tasks $T_i$ though the \textit{State Transformer}, i.e., $f_{s_i} =  ~$\textit{StateTransformer}$[f_{w_i}, f_{T_i}]$.

Then, we input $f_{s_i}$ into two Deep Q-networks, \textit{Q-network(w)} and \textit{Q-network(r)}, to predict Q values for each possible action $a_i$ at $s_i$, considering the benefit of workers $Q_w(s_i,a_i)$ and requesters $Q_r(s_i,a_i)$ separately. We use the aggregator/balancer to combine two benefits and generate the final action assigned to $w_i$. An explorer is also used to perform the trial-and-error actions. 


When $w_i$ is assigned one task, s/he can decide to complete or skip it. If $w_i$ sees a sorted list of tasks, we assume that workers follow a \emph{cascade model}\cite{click_model} to look through the task list and complete the first interesting task. The feedback is the completed task and the uncompleted tasks suggested to $w_i$.


Since the reward definitions are different in MDP(w) and MDP(r), we use two \textit{feedback transformers} to quantify the workers' feedback. As we said before, we explicitly predict transition probabilities and future states to ensure stable convergence and real-time behavior. Two \textit{future state predictors} are utilized for \textit{Q-Network(w)} and \textit{Q-Network(r)} separately, based on the historical statistics. 


If the action is to assign a task, we can store one transition $(s_i, a_i, r_i, s_{i+1})$ ($a_i$ is the assigned task) into the memory.
When the action is to recommend a list of tasks, 
the feedback includes the completed task and the uncompleted (suggested) tasks.
Thus, we store the successful transition $(s_i, a_i, r_i, s_{i+1})$ where $a_i$ is the completed task, and the failed transitions $(s_i, a_i, 0, s_{i+1})$ where $a_i$ is an uncompleted task. Each time we store one more transition into the memory, we use \textit{learner}s to update the parameters of two Q-networks, obtain a good estimation of $Q_w(s_i,a_i)$ and $Q_r(s_i,a_i)$ and derive the optimal policy $\pi$.
In the following sections, we will introduce the parts of the system in detail.  


\begin{figure*}[t!]
 \begin{minipage}[t]{0.72\linewidth}
 \centering

   \includegraphics[width=\textwidth]{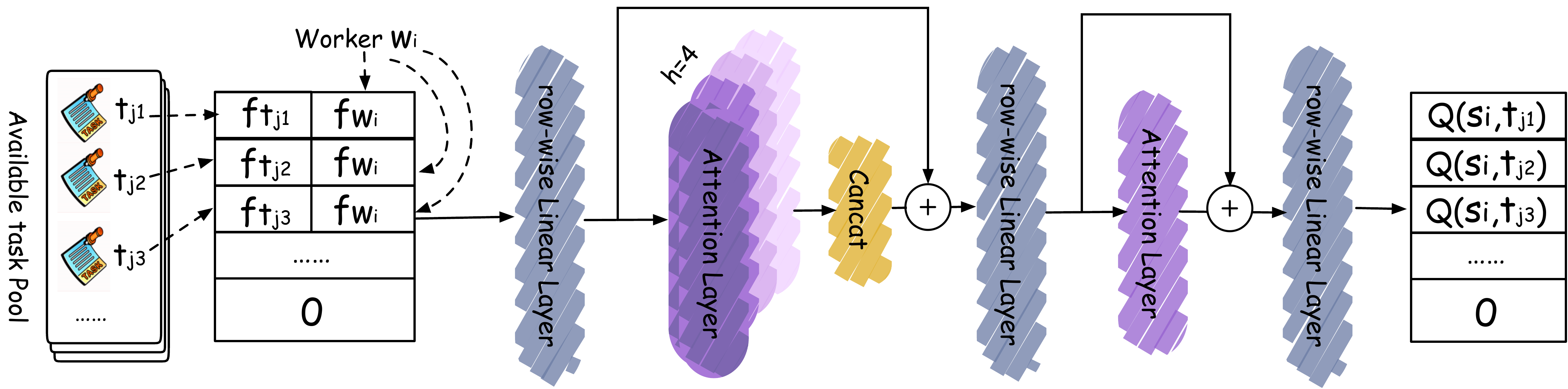}
     \caption{Q Network}
  \label{fig:Q_Network}
\end{minipage}
\begin{minipage}[t]{0.26\linewidth}
 \centering
   \includegraphics[width=0.5\textwidth]{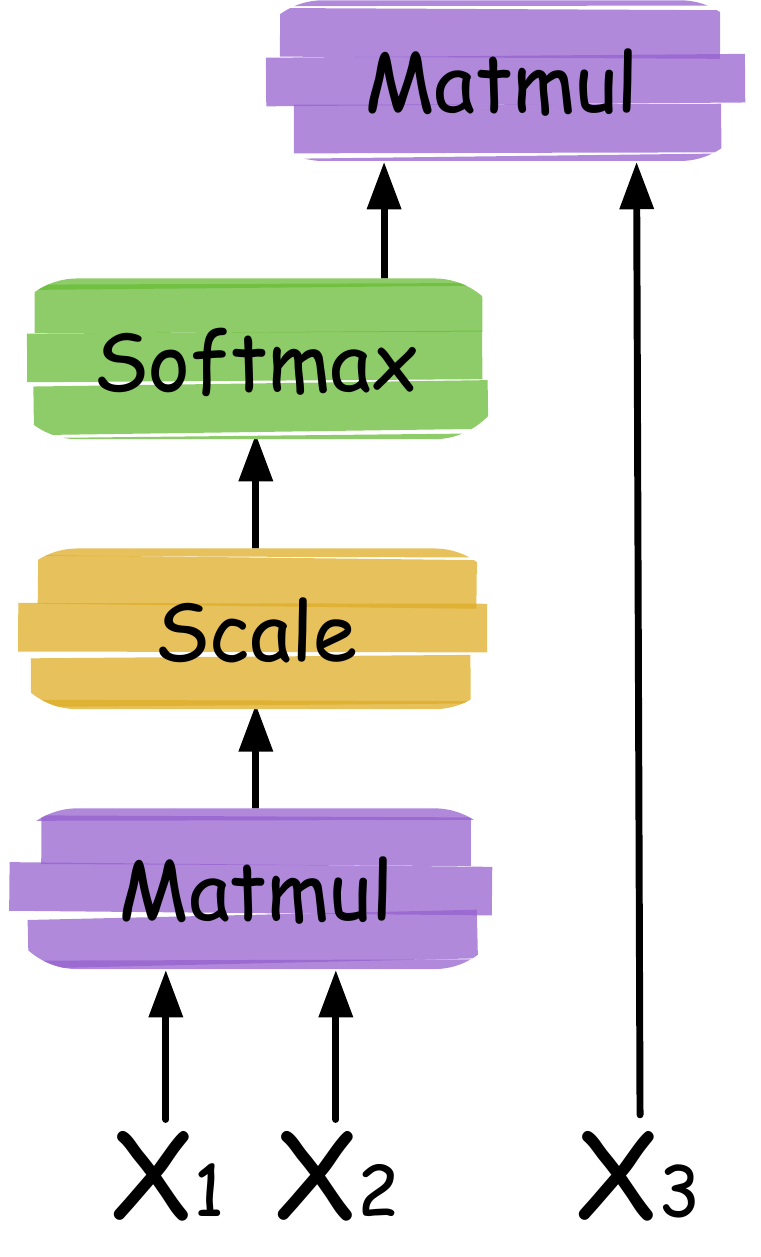}
   \caption{One Attention Layer}
  \label{fig:Attention_Layer} 
\end{minipage}  
\end{figure*}

\section{Modules for MDP(w)}\label{sec:method_ben_workers}

\subsection{Feature Construction} \label{sec:worker_feature_construction}

\subsubsection{{\bf Feature of a Task} $t_j$}
According to previous  studies\cite{kaufmann2011more}, the top-3 motivations of workers in crowdsourcing are the remuneration, the task autonomy and the skill variety. 
Task autonomy is the degree of freedom given to the worker for completing this task. Skill variety is the diversity of skills that are needed for solving and fit with the skill set of the worker. 

Thus, we construct the task features using \textit{award}, \textit{category} and \textit{domain}, which correspond to the top-3 three motives.
We use one-hot encoding to transform category and domain which are categorical attributes. 
Award is a continuous attribute which needs to be discretized. 
Then, we can concatenate them together to obtain the feature vector of task $t_j$. 


\subsubsection{{\bf Feature of a Worker} $w_i$}

In general, the features of a worker should be determined by the distribution of recently completed tasks by him/her (e.g., in the last week or month). This information can be used to  model the probability of a worker to complete a task in the near future.

\subsection{State Transformer and Q Network} \label{sec:worker_q_network}

\subsubsection{\bf Challenges}
We define the state $s_i$ to be composed of the set of available tasks $\{T_i\}$ and the worker $w_i$ at timestamp $i$.
However, it is hard to represent the set of available tasks. First of all, tasks are dynamic and their number is not fixed. We need to design a model can process input of any size. Secondly, the model should be {\em  permutation invariant} (i.e, it should not be affected by the order of tasks). 
Simple forward neural networks violate both requirements. Methods like LSTM\cite{LSTM} or GRU\cite{GRU} that process a variable-length sequences of data, are relative sensitive to the order. 

Some approaches in  recommender systems based on DQN \cite{zheng2018drn, zhao2018recommendations}  input the features of each task and  worker into a forward neural network independently to estimate the value of each task. However, they ignore the relationship among all available tasks. The value of a task is the same no matter which other tasks are available. This is not true in our setup because tasks are `competitive' and influence the value of other tasks.
Based on the above reasons, we need to design a novel representation for a set of available tasks. 

\subsubsection{\bf Design}
Inspired by \cite{deepset} and \cite{setTransformer}, we design our \textit{State Transformer} and {\it Q-Network} to obtain the state $s_i$ and values of each available task $Q(s_i,t_j)$, as
shown in Fig.~\ref{fig:Q_Network}. Firstly, we concatenate the features of each task $f_{t_{j*}}$ in the pool of available tasks with the feature of the worker $f_{w_i}$. To fix the length, we set the maximum value of an available task $\text{max}_T$ and use zero padding, i.e., add zeros to the end of $f_{s_i}$ and set its dimension to $[\text{max}_T, |f_{t_{j*}}|+|f_{w_i}|]$. 

Then we use row-wise Linear Layers and (multi-head) Attention Layers to project $f_{s_i}$ into $Q$ values, which keeps permutation-invariance. 
Row-wise Linear Layer is a row-wise feedforward layer which processes each row independently and identically. It calculates  function 
$$
\text{rFF}(X) = \text{relu}(XW+b)
$$
where $X$ is the input, $W$ and $b$ are the learnable parameters and relu is an activation function. 

The structure of the Attention Layer is shown in Fig.~\ref{fig:Attention_Layer}. Its input are three matrices $X_1 , X_2, X_3$ and it calculates 
$$
\text{Att}(X_1 , X_2, X_3) = \text{softmax}(\frac{X_1 X_2^{T}}{\sqrt{d}})X_3.
$$
The pairwise dot product $X_1 X_2^{T}$ measures how similar each row in $X_1$ and $X_2$ is, with a scaling factor of $\frac{1}{\sqrt{d}}$ and softmax function. The output is a weighed sum of $X_3$. Multi-head Attention Layer is proposed in \cite{attention}. It projects $X_1 , X_2, X_3$ into $h$ different matrices. The attention function Att is applied to each of the $h$ projections. The output is a linear transformation of the concatenation of all attention outputs. 
$$
\begin{aligned} 
&\text{MultiHead}(X_1 , X_2, X_3)  = \text{Concat}(\text{head}_{1}, ...,\text{head}_{h})W^{O} \\
&\text{  where  }\text{head}_{i}= \text{Att}(X_1 W^{X_1}_{i} , X_2 W^{X_2}_{i}, X_3 W^{X_3}_{i})
\end{aligned}
$$
We have to learn the parameters $\{W^{X_1}_{i} , W^{X_2}_{i}, W^{X_3}_{i}\}_{i=1}^{h}$ and $W^{O}$. Here we use multi-head Self-Attention layers, where $X_1=X_2=X_3=X$. When $X \in \mathbb{R}^{n \times d}$, a typical choice for the dimension of $W^{X}_{i}$ (resp. $W^{O}$) is $n \times \frac{d}{h}$ (resp. $n \times d$).

We can prove that row-wise Linear Layer and multi-head Self-Attention Layers are both permutation-invariant. The stack of these layers are also permutation-invariant. Please see the Appendix 
for details. 

We now summarize the design of our Q-network. Each row in the input $f_{s_i}$  is the pair of features of $t_j$ and $w_i$. The first two rFF layers are used to transform the task-worker features into high-dimensional features. Next, we use the multi-head self-attention layer to compute the pairwise interaction of different task-worker features in the set. Adding to the original features a rFF layer helps keeping the network stable. Thirdly, we use a self-attention layer again, which gives the Q-network the ability to compute pairwise as well as higher-order interactions among the elements in the set. The final rFF layer reduces the feature of each element into one value, representing $Q(s_i, t_j)$. 
Because of permutation-invariance, no matter the order of $t_j$, $Q(s_i, t_j)$ is the same. Besides, $Q(s_i, t_j)$ is decided by not only the pair of $w_i$ and $t_j$, also the other available tasks $t_{j'} \in T_{i}$.

\subsection{Action $\mathcal{A}$, Feedback and Reward $\mathcal{R}$} 

The workers of a crowdsourcing platform aim at achieving a good experience. Payment-driven workers aim at finding high award per unit of time tasks while interest-driven workers hope to answer tasks that match their interest. Mixed-interest workers decide by balancing these factors. Our goal is to help them in finding tasks interesting to them as soon as possible, i.e., at maximizing the completion rate of recommended tasks.

If the agent is to assign one task, it selects the action $a_i=t_j$ with the maximum $Q(s_i, t_j)$. We assume workers follow a \emph{cascade model} to look through the list of tasks, so if the agent recommends a task list, the action 
is $\sigma(T_i)=\{t_{j_1},t_{j_2},...\}$ where $t_{j*}$'s are  ranked in descending order of $Q(s_i, t_{j*})$. 

As for the feedback and reward, the feedback is completed or skipped when the action is one task. Thus, the immediate reward is $1$ if the worker completes the task or $0$ if the worker rejects it. 
When the action is a list of $k$ tasks,
the immediate reward is $1$ if the worker finishes one of the tasks or $0$ if the worker never finishes any of them. 

\subsection{Future State, Memory Storer, and Learner} \label{sec:future_state_w}

\begin{figure*}[t!]
\centering  
   \includegraphics[width=0.85\linewidth]{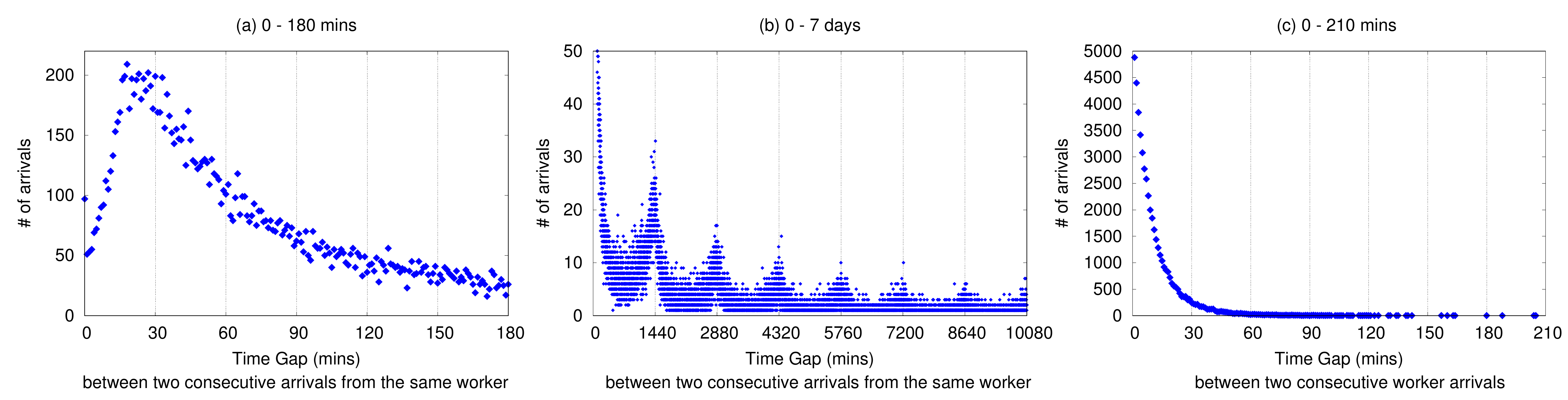}
   \caption{Time Gap between Two Consecutive Arrivals of Workers}
  \label{fig:time_gap_worker_request} 
\end{figure*}

\subsubsection{\bf Challenges}
The future state $s_{i+1}$ is the timestamp when the same worker $w_i$ comes again. Thus, the time of receiving $r_i$ and the future state $s_{i+1}$ is different. Besides, it may take a long time for the same worker to come again (the median value of the time gap is one day in our data) and for the transition $(s_i, a_i, r_i, s_{i+1})$ to be stored. Because the parameters in {\it Q-network(w)} are shared by all  workers, not knowing the latest transitions may harm the performance. 

Therefore, we design a {\it predictor(w)} to predict the transition probability $Pr(s_{i+1}|s_i,a_i,r_i)$ and the feature of the future state $f_{s_{i+1}}$ after we obtain the feedback and reward $r_i$ for $(s_i,a_i)$. This helps our framework to satisfy the requirement of handling online changes and achieving real-time interaction. 


\subsubsection{\bf Design}
First of all, the worker feature $f_{w_i}$, i.e., the distribution of recently completed tasks, needs to be updated by $r_i$. Based on the MDP(w) definition, $w_{i+1}=w_i$ and the worker feature $f_{w_{i+1}}$ at $s_{i+1}$ is the updated feature $f_{w_{i}}$.


Secondly, we consider $T_{i+1}$ and its feature $f_{T_{i+1}}$ at $s_{i+1}$. The change between $T_i$ and $T_{i+1}$ comes mainly from the expired tasks. We need to check whether $t_j \in T_{i}$ has expired at $\text{Time}_{i+1}$ (i.e., the happening time of $s_{i+1}$) and remove expired tasks from $T_{i+1}$. 

$\text{Time}_{i+1}$ is stochastic and we need to learn its distribution from the environment. From the history, we find that there is a pattern of the same worker arrivals, i.e., a worker comes again within a short time, or comes again after 1 day, 2 days, etc. up to one week later (see the distribution of the time gap between two consecutive arrivals from the same worker in Fig.~\ref{fig:time_gap_worker_request}(a) and \ref{fig:time_gap_worker_request}(b)). 
To capture the pattern, we maintain a function $\phi(g)$, where $g$ is the time gap, and $\phi(g = \text{CurrentTime} - \text{TimeOfLastArrival}_{w})$ is the probability whether the worker $w$ comes again currently. We set $g \in [1,10080]$ minutes since the probability of $\phi(g)>0, g>10080$ is small and can be ignored. Note that $\phi(g)$ is initialized by the history and iterative updated when we have a new sample. 

Finally the distribution of $\text{Time}_{i+1}$ is $\text{Time}_{i} + \phi(g)$, $g \in [1,10080]$. Given a possible $\text{Time}_{i+1}$, {\it predictor(w)} checks whether tasks are expired and generates $s_{i+1}$ and $f_{s_{i+1}}$.


For {\it learner(w)}, we use the method introduced in Sec.~\ref{sec:RL_basic} to update the parameters of {\it Q-Network(w)} by transitions stored in the memory. Our loss function can be written as 
\begin{equation} \label{eq:q_network_w_lf}
\begin{aligned}
L(\theta)  &= \mathbb{E}_{\{(s_i,a_i,r_i)\}}[(y_i-Q(s_i,a_i;\theta))^{2}] \\
y_i  &=  r_i+ \gamma \sum_{g} Pr(s_{i+1}|g) \max_{a_{i+1}} Q(s_{i+1},a_{i+1};\theta)
\end{aligned}
\end{equation}
where $Pr(s_{i+1}|g) = \phi(g)$ and $g \in [1,10080]$.
Actually, we do not need to calculate $\max_{a_{i+1}} Q(s_{i+1},a_{i+1};\theta)$ for all  possible $g$. The value $\max_{a_{i+1}}Q$ may change when a task $ t_{j'} \in T_i$ expires. Thus, the maximum times we compute $\max_{a_{i+1}}Q$ is $max_{T}$.

Here, we also use the double Q-learning algorithm\cite{double_q_learning} to avoid overestimating Q values. The algorithm uses another neural network $\widetilde{Q}$ with parameters $\widetilde{\theta}$, which has the same structure as the Q-Network $Q$, to select actions. The original Q-Network $Q$ with parameters $\theta$ is used to evaluate actions. That is:
$$ \small
y_i  =  r_i+ \gamma \sum_{g} Pr(s_{i+1}|g) \widetilde{Q} (s_{i+1}, \arg \max_{a_{i+1}} Q(s_{i+1},a_{i+1}|\theta)|\widetilde{\theta}).
$$
Parameters $\widetilde{\theta}$ are slowly copied from parameters $\theta$ during  learning. 

Accordingly, the gradient update is 
\begin{equation} \label{eq:q_network_w_uq} \small
\begin{aligned}
&\nabla_{\theta} L(\theta) = \mathbb{E}_{\{(s_i,a_i,r_i)\}}[ r_i + \gamma \sum_{g} Pr(s_{i+1}|g) \\
& \quad \quad \widetilde{Q} (s_{i+1}, \arg \max_{a_{i+1}} Q(s_{i+1},a_{i+1}|\theta)|\widetilde{\theta}) -  Q(s_i,a_i) ]\nabla_{\theta}Q(s_i,a_i).
\end{aligned}
\end{equation}
Prioritized experience replay \cite{prioritized_er} is used to learn efficiently.

\section{Modules for MDP(r)}\label{sec:method_ben_requesters}

\subsection{Feature Construction}

In addition to the features of tasks and workers introduced in Sec.~\ref{sec:worker_feature_construction}, we also  use the quality of workers $q_{w_i} \in [0,1]$ and the quality of tasks $q_{t_j} \in \mathbb{R}$ to predict the benefit of requesters.
We assume that we already know the the quality of workers from their worker answer history or the qualification tests with the ground truth. 
The quality of tasks is decided by all the workers who completed it.  
We assume that workers who come at timestamps $i \in I_{t_j}$, complete the task $t_j$. We use the Dixit-Stiglitz preference model\cite{dixit} to calculate task quality $q_{t_j}$ based on the law of {\em diminishing marginal utility}. That is:
\begin{equation} \label{eq:compute_task_quality}
q_{t_j} = (\sum_{i \in I_{t_j}} (q_{w_i})^p)^{1/p} ,p \geq 1.
\end{equation}
Note that the same worker can come several times at different timestamps. $p$ controls how much marginal utility we can get with one more worker.

Let us explain the above equation using two typical examples. The first is AMT, where each task has multiple independent micro-tasks and each micro-task is only allowed to be answered by one worker. The quality of mirco-tasks is equal to the quality of the answering worker. Since the micro-tasks are independent, the quality of the task is the sum of the qualities of the micro-tasks which comprise it, where $q_{t_j} = \sum_{i \in I_{t_j}} q_{w_i}$, $p=1$.
The second example is  competition-based crowdsourcing platforms, where tasks can be answered by many workers, but only one worker is selected to be awarded after the deadline. 
The quality should be defined as $q_{t_j} = \max_{i \in I_{t_j}} q_{w_i}$, i.e., $p$ is set to infinity.

\subsection{State Transformer and Q Network}
The {\it State Transformer} and the  {\it Q-Network} are as defined in Sec. ~\ref{sec:worker_q_network}; 
 we only need to add  the two dimensions ($q_{w_i}$ and $q_{t_j}$) to the input. 

\subsection{Action $\mathcal{A}$, Feedback and Reward $\mathcal{R}$}  

Same as before, the action $a_i = t_j$ with the maximum $Q_r(s_i, t_j)$ is recommended, if the agent assigns one task to $w_i$. To recommend a list, the action is $a_i =\sigma(T_i)=\{t_{j_1},t_{j_2},...\}$, where $t_{j*}$'s are ranked in descending order of $Q_r(s_i, t_{j*})$.

From the requester's perspective, the goal is to obtain the greatest possible quality of results before the deadline of tasks. Thus the immediate reward is 
$q_{t_j}^\text{new}-q_{t_j}^\text{old}$ if the worker is assigned to the task $t_j$ and finishes it. The reward is $0$ if the worker skips the task. When the action is to recommend a list of $k$ tasks, the immediate reward is $q_{t_{j*}}^\text{new}-q_{t_{j*}}^\text{old}$ if the worker selects the task $q_{t_{j*}}$ and completes it.  The reward is $0$ if the worker does not finish any task. 


\subsection{Future State, Memory Storer and Learner}
\label{sec:future_state_r}

\subsubsection{\bf Challenges}
Different from MDP(w), the next worker in MDP(r) arrives fast.
However, we find that when we use the real worker $w_{i+1}$ and $T_{i+1}$ to combine $s_{i+1}$, it is hard for Deep Q-network to converge. Varying next workers make diverse states and transitions sparse, leading to inaccurate estimation of transition probability and unstable convergence. Hence, we use the expectation of the next worker instead of the real next worker to train {\it Q-network(r)}. 

\subsubsection{\bf Design}
After we obtain the feedback and reward $r_i$ for $(s_i,a_i)$, the first thing is to update the worker feature $f_{w_i}$ when $r_i>0$. Besides, we also need to update the quality in the task feature $f_{t_j}$ which is completed. 

From the benefit of requesters, the qualities of tasks are influenced by all workers. Thus the future state $s_{i+1}$ happens when the next worker $w_{i+1}$ (no matter whether $w_{i+1}= w_i$) comes. Here the {\it future state predictor(r)} not only needs to estimate the next timestamp and check for expired tasks, but also has to predict the next worker.

We first explain how we predict $\text{Time}_{i+1}$. Fig.~\ref{fig:time_gap_worker_request}(c) shows the distribution of the time gap between two consecutive arrivals, no matter whether these two arrivals are from the same or different workers. It is a long-tail distribution, which means that workers come to the platform and complete tasks frequently. 
We also maintain a function $\varphi(g)$, where $g$ is the time gap, and $\varphi(g =\text{Time}_{i+1}-\text{Time}_{i})$ is the probability that a worker comes at $\text{Time}_{i+1}$ if the last worker comes at $\text{Time}_{i}$. We set $g \in [0,60]$ minutes because $99\%$ of time gaps in the history are smaller than 60 minutes. Same as $\phi(g)$, $\varphi(g)$ is also built from the history and iteratively updated at each new sample. Then the distribution of $\text{Time}_{i+1}$ is $\text{Time}_{i} + \varphi(g)$.

After we know $\text{Time}_{i+1}$, we compute the distribution of the coming workers. For each worker $w \in W^\text{old}$ who already came before, we know the feature of worker $f_w$ and the time gap between his/her last arrival time and $\text{Time}_{i+1}$ (i.e., $g_{w}=\text{Time}_{i+1} -\text{TimeOfLastArrival}_{w}$). From  function $\phi(g)$ defined in Sec.~\ref{sec:future_state_w}, we obtain  probability $\phi(g_{w})$. Besides, we also consider the probability that a new worker comes. From the history, we also maintain the rate of new workers $p_{\text{new}}$, and we use the average feature of old workers $\bar{f_w}$ to represent the feature of new workers. Finally, we normalize, integrate and obtain the probability for a coming worker $w$:
$$ \small
Pr(w_{i+1}=w) = 
\begin{cases}
(1-p_{\text{new}})\frac{\phi(g_{w})}{\sum_{w'\in W^\text{old}}\phi(g_{w'})} & \text{ when } w \in W^\text{old} \\
p_{\text{new}}  & \text{ when } w \text{ is new} \\
\end{cases}
$$
Given  $g$ and $w_{i+1}$, we use the  method described in Sec.\ref{sec:future_state_w} to calculate  $T_{i+1}$ and $s_{i+1}$. 

For {\it learner(r)}, our loss function is
\begin{equation}\label{eq:q_network_r_lf} \small
\begin{aligned}
L(\theta)  &= \mathbb{E}_{\{(s_i,a_i,r_i)\}}[(y_i-Q(s_i,a_i;\theta))^{2}] \\
y_i  &=  r_i+ \gamma \sum_{g} \sum_{w_{i+1}} Pr(s_{i+1}|g, w_{i+1}) \\
&\quad \quad \quad\quad\quad\quad \quad\quad\quad \widetilde{Q} (s_{i+1}, \arg \max_{a_{i+1}} Q(s_{i+1},a_{i+1}|\theta)|\widetilde{\theta})
\end{aligned}
\end{equation}
where $Pr(s_{i+1}|g, w_{i+1}) = \varphi(g)Pr(w_{i+1}|g)$ and $g \in [0,60]$ while $w_{i+1} \in W^\text{old}$ or $w_{i+1}$ is new.

Accordingly, the gradient update is 
\begin{equation}\label{eq:q_network_r_uq} \small
\begin{aligned}
&\nabla_{\theta} L(\theta) = \mathbb{E}_{\{(s_i,a_i,r_i)\}}[ r_i + \gamma \sum_{g} \sum_{w_{i+1}} Pr(s_{i+1}|g,w_{i+1}) \\
& \quad \quad \widetilde{Q} (s_{i+1}, \arg \max_{a_{i+1}} Q(s_{i+1},a_{i+1}|\theta)|\widetilde{\theta}) -  Q(s_i,a_i) ]\nabla_{\theta}Q(s_i,a_i).
\end{aligned}
\end{equation}

However, computing $\widetilde{Q} (s_{i+1}, \arg \max_{a_{i+1}} Q(s_{i+1},a_{i+1}))$ for all possible $g$ and $w_{i+1}$ may take a long time. Here are two methods to speed this up. One method is to limit the number of possible workers. We can set a threshold to disregard workers with low coming probability. Another method is to use the expectation of the feature of all possible $w_{i+1}$ instead of computing them. The expectation of the feature of the next worker is $\bar{f}_{w_{i+1}} = \sum_{w_{i+1}} Pr(w_{i+1}|g) f_{w_{i+1}}$, the expectation of future state feature is $\bar{f}_{s_{i+1}} = [\bar{f}_{w_{i+1}},f_{T_{i+1}}]$ and the loss function and updating equation are given by Eq.~\ref{eq:q_network_w_lf} and Eq.~\ref{eq:q_network_w_uq}, respectively.

\section{Integration of MDP(w) and MDP(r)}\label{sec:integration}

\subsection{Aggregator and Balancer}
The profit model of commercial platforms (i.e., AMT) is to charge a percentage of the award given to  workers who finish their tasks.
Thus, the platform aims at attracting more workers and requesters. 
To achieve this goal the  platform should satisfy 
workers and requesters simultaneously. 

Based on {\it Q-network(w)} and {\it Q-network(r)}, we obtain the Q values $Q_w(s_i,t_j)$ and $Q_r(s_i,t_j)$ for each available task $t_j$ separately. When we recommend $t_j$ at $s_i$, 
$Q_w(s_i,t_j)$ represents the predicted Q value for the worker $w_i$, while $Q_r(s_i,t_j)$ represents the predicted Q value for the currently available tasks $T_i$.  
We use weighted sum to balance them to a single predicted Q value $Q(s_i,t_j) = w Q_w(s_i,t_j)+ (1-w)Q_r(s_i,t_j)$.

Same as before, we either select the task $a_i = t_j$ with the maximum $Q(s_i,t_j)$, or arrange and show a list of tasks $a_i = \sigma(T_i) = \{t_{j_1}, t_{j_2} , ..., t_{j_n}\}$ 
in descending $Q(s_i,a_{j*})$ order. 

\subsection{Explorer}

The most straightforward strategy to conduct exploration in reinforcement learning is $\epsilon$-greedy \cite{epsilon_greedy}. 
This approach randomly selects a task or sorts tasks with a probability of $\epsilon$, or follows $Q(s_i,t_j)$ to recommend a list of tasks with probability $1-\epsilon$.
This is suitable for recommending one task but does not perform well in recommending a list of tasks because it is too random. 
Instead of ignoring $Q(s_i,t_j)$ totally, we add a random value $v$ into $Q(s_i,t_j)$ with a probability of $\epsilon$. We generate $v$ as a normal distribution where the mean is zero and the standard deviation is the same as that of the current Q values ($Q(s_i,t_j), \forall t_j \in \{T_i\}$). Besides, we also use a decay factor to multiply the standard deviation, in order to reduce randomness when the Q-network is relatively mature. 

\section{Experiments}\label{sec:experiments}

\subsection{Experimental Settings}\label{sec:exp}

\subsubsection{Dataset}
\begin{figure}[t!]
\centering  
   \includegraphics[width=\linewidth]{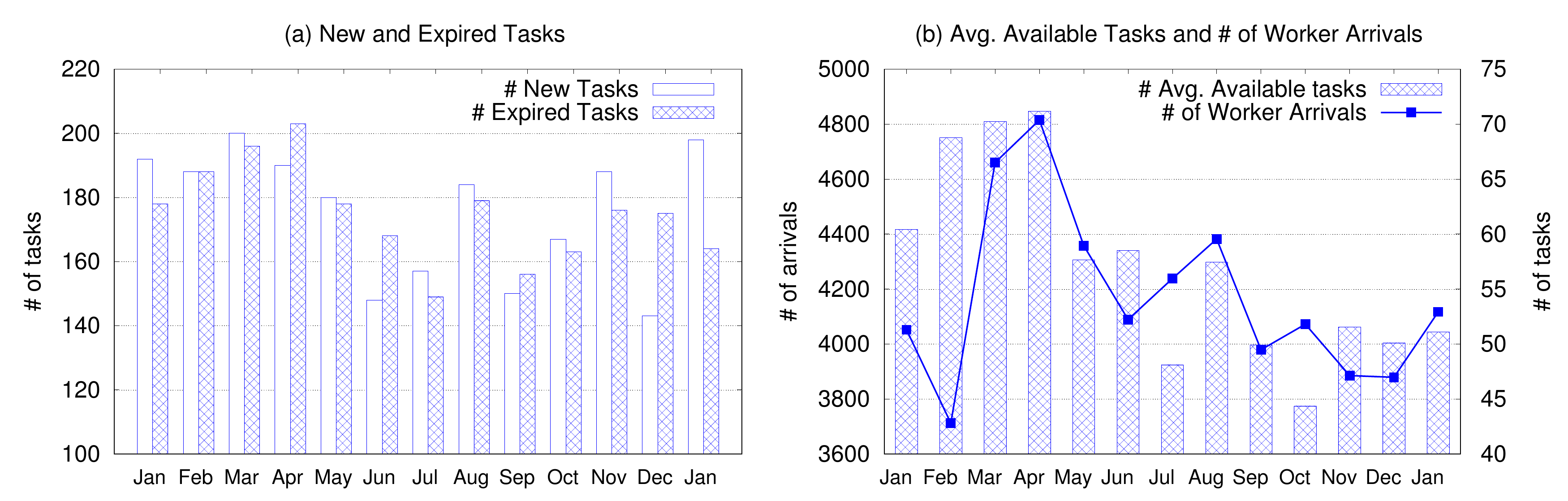}
   \caption{New/Expired/Available Tasks and Worker Arrivals}
  \label{fig:n_new_expired_task_all} 
\end{figure}

We conduct  experiments on a real dataset collected from the commercial crowdsourced platform CrowdSpring \cite{CrowdSpring}.
This platform helps requesters publish tasks to obtain high-quality custom logos, names, designs, etc. 
Most of the tasks are public, i.e.,  we can see all the information including start date and deadline, category, sub-category, domain and the relationship of workers who completed it. 
We use a web crawler to obtain all the information about public tasks ranging from Jan 2018 to Jan 2019. 
There are totally 2285 tasks created and 2273 tasks expired.
There are about 1700 active workers during the entire process.
We show the number of new and expired tasks per month in Fig.~\ref{fig:n_new_expired_task_all}(a), which are around 180. Besides, Fig.~\ref{fig:n_new_expired_task_all}(b) shows the number of arrivals of workers per month and how many available tasks they can select to complete. There are about 4200 arrivals of workers per month. When a worker comes, s/he can see 56.8 available tasks on average. 

We also generated a synthetic dataset, simulating the real dataset using factors considered in \cite{RLBipartiteGraphMatching}. We consider the arriving density of workers, the distribution of qualities of workers and scalability.


\subsubsection{Evaluation Measures}
Depending on whether the agent recommends one task or a list of tasks, and considering the benefit of workers or requesters, we use the following measures to evaluate the performance of methods. 

\stitle{For the benefit of workers:}
\begin{itemize}
	\item Worker Completion Rate (\textbf{CR}). At timestamp $i$ the worker $w_i$ comes, the agent recommends a task $t_j$. We compute the cumulative number of completions rate where $y_{ij} = 1$ means that the task is completed and $y_{ij} = 0$ means that the task is skipped. 
	\begin{equation}
		CR =  \frac{ \sum_{i} y_{ij} }{ \text{number of total timestamps}}    
	\end{equation}

	\item \textbf{nDCG-CR}. Instead of one task, the agent recommends a list of tasks. We apply the standard Normalized Discount Cumulative Gain proposed in \cite{nDCG} to evaluate the success of the recommended list $L=\{t_{j_1},t_{j_2},...,t_{j_{n_i}}\}$ for all available tasks at timestamp $i$. $r$ is the rank position of tasks in the list, $n_i$ is the number of available tasks. We assume that $w_i$ looks through the tasks in order and completes the first task $t_{j_r}$ s/he is interested in. $y_{ij_r}=1$ indicates that $t_{ij_r}$ is completed; all other $y_{ij_{r'}}$ are  $0$. 
	\begin{equation}
		nDCG-CR = \frac {\sum_{i}  \sum_{r=1}^{n_i} \frac{1}{log(1+r)} y_{ij_r}}{ \text{number of total timestamps}}
	\end{equation}
	\item Top-$k$ Completion Rate (\textbf{kCR}). We limit the length of the list to $k$, i.e., the agent recommends $k$ tasks $\{t_{j_1},t_{j_2},...,t_{j_k}\}$ for the worker $w_i$. We assume that $k$ tasks also have an order and that $w_i$ looks through the tasks in order and completes the first interesting task $t_{j_r}$.  
	\begin{equation}
		kCR = \frac {\sum_{i}  \sum_{r=1}^{k} \frac{1}{log(1+r)} y_{ij_r}}{ \text{number of total timestamps}}  
	\end{equation}

\end{itemize}

\stitle{For the benefit of requesters:}
\begin{itemize}
	\item Task Quality Gain (\textbf{QG}). At timestamp $i$, worker $w_i$ comes and the agent recommends a task $t_j$. We compute the cumulative gain of the qualities of tasks. If the task is skipped, $g_{ij} = 0$. Otherwise, $g_{ij}$ is the difference of the task quality $q_{t_j}$ before and after $w_i$ finishes $t_j$.
	\begin{equation}
		QG =  \sum_{i} g_{ij} =  \sum_{i} q^{\text{new}}_{t_j} - q^{\text{old}}_{t_j}  
	\end{equation}
	\item \textbf{nDCG-QG}. Same as \textbf{nDCG-CR}, we apply nDCG to give different weights for rank positions of tasks. $y_{ij_r}$ indicates whether $t_{j_r}$ is completed, and $g_{ij_r}$ is the gain in the quality of $t_{j_r}$.
	\begin{equation}
		nDCG-QG = \sum_{i} \sum_{r=1}^{n_i} \frac{1}{log(1+r)} y_{ij_r} g_{ij_r}
	\end{equation}
	\item Top-$k$ Task Quality Gain (\textbf{kQG}). Similarly, we limit the recommended list into $k$ tasks $\{t_{j_1},t_{j_2},...,t_{j_k}\}$ for the worker $w_i$. 
	\begin{equation}
		kQG =  \sum_{i} \sum_{r=1}^{k} \frac{1}{log(1+r)} y_{ij_r} g_{ij_r}  
	\end{equation}

\end{itemize}

\subsubsection{Competitors}
We compared our approach with five alternative methods. The worker and task features of all these methods are updated in real-time. The methods using supervised learning (Taskrec(PMF)/Greedy+Cosine Similarity/Greedy+Neural Network) predict the completion probability and the quality gain of tasks and select one available task or sort the available tasks based on predicted values. The parameters of the models are updated at the end of each day. For the reinforcement learning methods (LinUCB/DDQN), the parameters are updated in real-time after one recommendation.

\begin{itemize}
	\item \textbf{Random}. For each worker arrival, one available task is picked randomly, or a list of tasks is randomly sorted and recommended. 

	\item \textbf{Taskrec (PMF)}. Taskrec\cite{taskrec} is a task recommendation framework for crowdsourcing systems based on unified probabilistic matrix factorization. Taskrec builds the relationship between the worker-task, worker-category and task-category matrices and predicts the worker completion probability. It only considers the benefit of workers.

	\item \textbf{SpatialUCB/LinUCB}. SpatialUCB\cite{spatialUCB} adapts the Linear Upper Confidence Bound\cite{linUCB} algorithm in online spatial task assignment. We adapt SpatialUCB in our setting by replacing the worker and task features. SpatialUCB selects one available task or sorts the available tasks according to the estimated upper confidence bound of the potential reward. For the benefit of requesters, we add the quality of workers and tasks as features and then predict the gain quality of the tasks.  

	\item \textbf{Greedy+Cosine Similarity}. We regard the cosine similarity between the worker feature and task feature as the completion rate, and select or sort tasks greedily according to the completion rate. For the benefit of requesters, we use the actual value of the quality gain by multiplying the completion probability of each task to pick or rank the available tasks.  

	\item \textbf{Greedy+Neural Network}. We input the worker and task features into a neural network of two hidden-layers to predict the completion rate. For the benefit of requesters, we add the quality of workers and tasks as features and then predict the gain quality of the tasks.  

	\item \textbf{DDQN}. Double Deep Q-Network is our proposed framework, In the first two experiments, we use a version of DDQN that only considers the benefit of workers or requesters when comparing it with the other approaches. 

\end{itemize}

\begin{figure*}[t!]
 \begin{minipage}[c]{0.65\linewidth}
   \centering  
   \includegraphics[width=\linewidth]{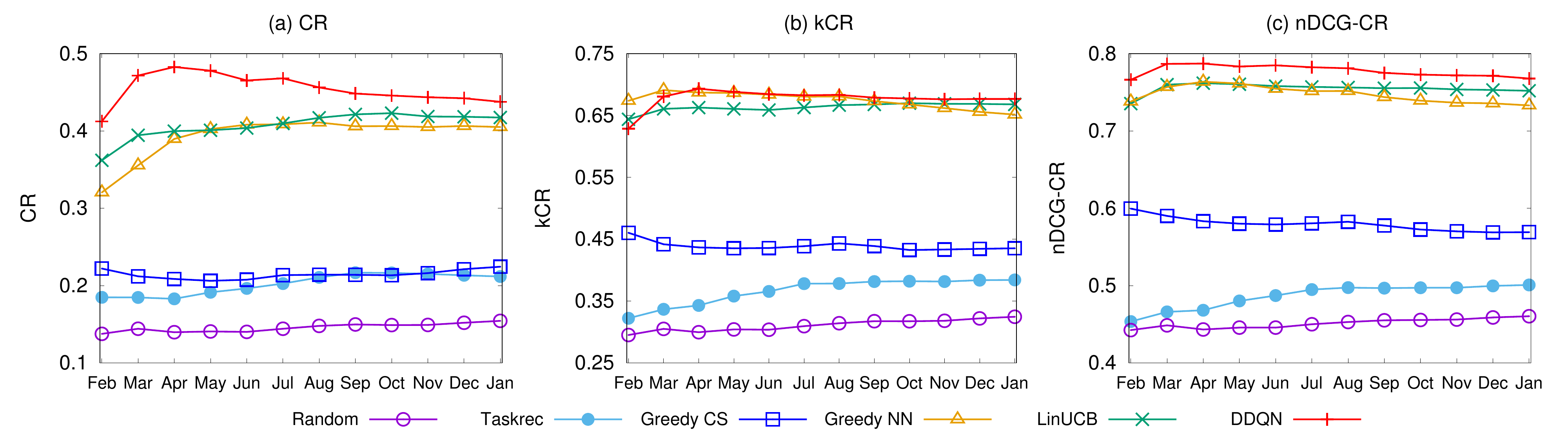}
\end{minipage}
\begin{minipage}[c]{0.3\linewidth} 
\centering  
\scalebox{0.9}{
\begin{tabular}{|c|c|c|c|}
\hline
       & CR 				& kCR 				& nDCG-CR \\ \hline
Random    & 0.154   		& 0.325    			&  0.460       \\ \hline
Taskrec   & 0.212  			& 0.384    			&  0.501       \\ \hline
Greedy CS & 0.224   		& 0.435    			&  0.569       \\ \hline
Greedy NN & 0.405   		& 0.651    			&  0.733       \\ \hline
LinUCB    & 0.417   		& 0.668    			&  0.752       \\ \hline
DDQN      & {\bf 0.438}   	& {\bf 0.677}    	& {\bf 0.768}       \\ \hline
\end{tabular}
}
\end{minipage}    
\caption{Benefits of Workers}
\label{fig:worker_benefit} 
\end{figure*}

\begin{figure*}[t!]
 \begin{minipage}[c]{0.65\linewidth}
   \centering  
   \includegraphics[width=\linewidth]{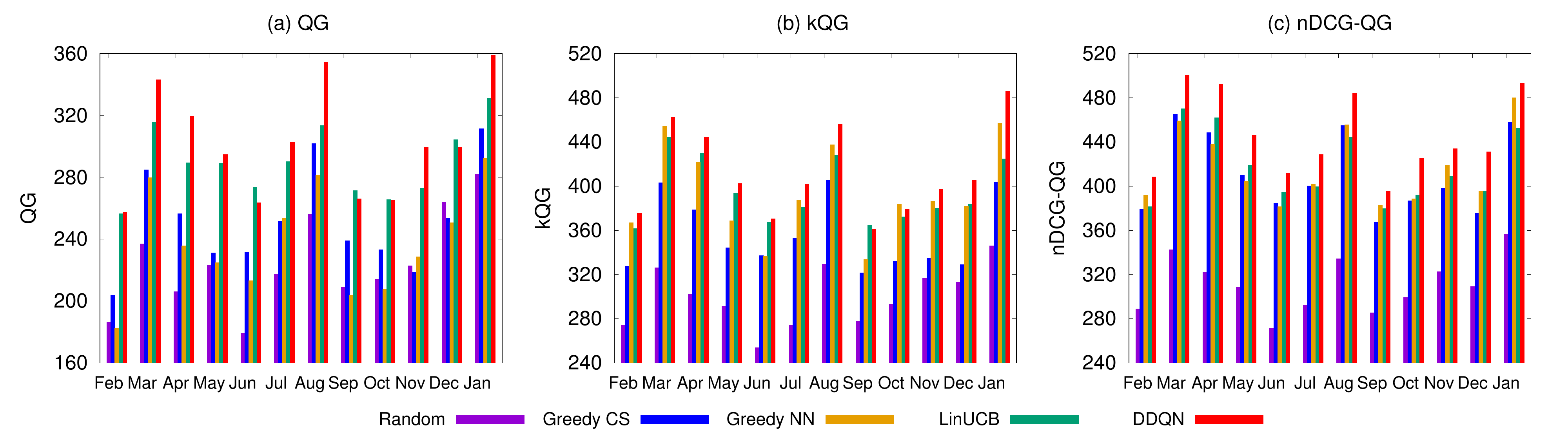}
\end{minipage}
\begin{minipage}[c]{0.3\linewidth} 
\centering  
\scalebox{0.9}{
\begin{tabular}{|c|c|c|c|}
\hline
       & QR 				& kQR				& nDCG-QR \\ \hline
Random    & 2697.96   		& 3598.05    			&  3733.52       \\ \hline
Greedy CS & 3017.46   		& 4269.64   			&  4929.46      \\ \hline
Greedy NN & 2854.58   		& 4716.83    			&  4998.76       \\ \hline
LinUCB    & 3474.04   		& 4731.97    			&  4999.67       \\ \hline
DDQN      & {\bf 3625.34}   	& {\bf 4943.29}    	& {\bf 5350.98}       \\ \hline
\end{tabular}
}
\end{minipage}  
\caption{Benefits of Requesters}
\label{fig:requester_benefit} 
\end{figure*}

\begin{figure*}[t!]
\begin{minipage}[c]{0.75\linewidth}
\centering  
\includegraphics[width=\linewidth]{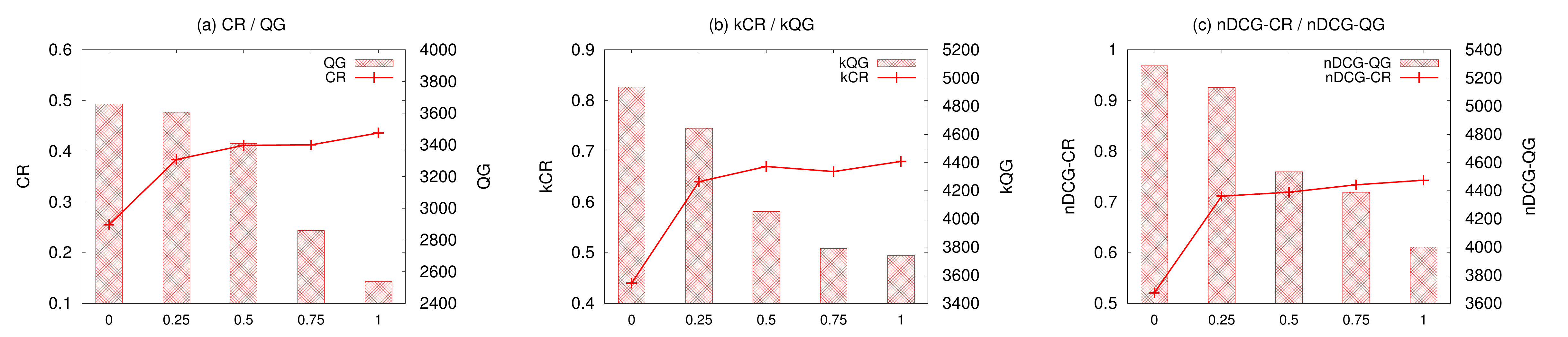}
\caption{Balance of Benefits}
\label{fig:balance_benefit} 
\end{minipage} 
\begin{minipage}[c]{0.22\linewidth} 
\centering  
\begin{tabular}{|c|c|}
\hline
           & Time (Sec)   \\ \hline
Taskrec         & 3.193	      \\ \hline
Greedy NN       & 7.476   		  \\ \hline
LinUCB           & 0.073   		  \\ \hline
DDQN             & 0.042       \\ \hline
\end{tabular}
\captionof{table}{Efficiency}
\label{table:efficiency} 
\end{minipage} 
\end{figure*}

\subsection{Experimental Results (real dataset)}

\subsubsection{Implementation details}

The dataset is static and records the cases where workers complete certain tasks. It does not include any information about  tasks for which  workers were not interested. Since the number of available tasks at a time is $\sim50$, we assume that a  worker who arrives looks through all available tasks and completes one which he/she finds interesting, so the remaining tasks that workers see but skip are considered not interesting.

We order the dataset, i.e., creation of tasks, expiration of tasks and arrival of workers by time. We use the data in the first month (Jan 2018) to initialize the feature of workers and tasks and the learning model. Then, we simulate the process that a worker comes, a task is created or expires as time goes by. The entire process runs from Feb 2018 to Jan 2019. To solve the cold-start problem of new workers, we also use the first five tasks they completed to initialize their features. 

We set $p=2$ to compute the quality of tasks by Eq.~\ref{eq:compute_task_quality}. 
The dimension of output features in each layer of Q-Network is set to $128$. 
The buffer size for DDQN is $1000$ and we copy parameters $\widetilde{\theta}$ from $\theta$ after each $100$ iterations. The learning rate is $0.001$ and the batch size is $64$. We set the discount factor $\gamma = 0.5$ for the benefit of requesters and $\gamma = 0.3$ for workers. 
To do the exploration, we set the initial $\epsilon=0.9$, and increase it until $\epsilon=0.98$ for assigning a task. To recommend the task list, $\epsilon$ is always $0.9$, and the decay factor for standard deviations is set as $1$ initially and decreases into $0.1$ with further learning. We use Pytorch to implement all the algorithms and used a GeForce GTX 1080 Ti GPU. 

\subsubsection{Considering the benefit of workers}
We show QR, kQR and nDCG-QR for each method at the end of each month in Fig.~\ref{fig:worker_benefit}.
Random performs the worst since it never predicts the worker completion probability. 
The reason behind the bad performance of Taskrec is that it only uses the category of tasks and workers and ignores the domain or award information.   
Because of the simple model to compute the similarity of tasks for a certain worker, Greedy CS also performs badly.
Greedy NN uses the neural network to predict the relationship between tasks and workers, and updates the parameters every day. However, it only considers the immediate reward. Thus it performs worse than LinUCB and DDQN.
LinUCB utilizes all information of features of workers and tasks, estimates the upper confidence bound of the reward and updates parameters after each worker feedback.
So its performance is second to DDQN.    
Our proposed model, DDQN, not only uses the neural network to model the complex relationship between workers and tasks, but also predicts the immediate and future reward and updates the parameters after each worker feedback. Therefore, DDQN outperforms all competitors.

The table lists the final value of CR, kCR and nDCG-CR of each method; our approach is round 2\% better than other models.

\subsubsection{Considering the benefit of requesters}
We show the separate quality gain of tasks in each month in Fig.~\ref{fig:requester_benefit}. Note that the gain is not 
consistently increasing but it is related to the number of worker requests at each month in Fig.~\ref{fig:n_new_expired_task_all}(b).  
The random method again performs the worst. 
Although we give the real value of the quality gain of each task, Greedy CS still cannot recommend tasks with the high gain which are completed by workers. 
Greedy NN and LinUCB perform similarly (in kQR and nDCG-QR). 
Greedy NN achieves a better estimation than LinUCB when
aggregating the quality gain and completion rate of each task,
while LinUCB could update the model more timely. 
Still, the performance of DDQN is the best because it utilizes the nonlinear and complex Q-network to approximate, predict and integrate the gain and completion rate of tasks in the long term. 

The table lists the final value of QR, kQR and nDCG-QR of each method; our method is at least 4.3\% better than its competitors.

\subsubsection{Balance of benefits}
We integrate the two benefits of workers and requesters using the weighed sum model $Q(s_i,t_j) = w Q_w(s_i,t_j)+ (1-w)Q_r(s_i,t_j)$ and show the result in Fig.~\ref{fig:balance_benefit}. We test the cases of $w=0, 0.25, 0.5, 0.75$ and $1.0$. 
From the trend of CR and QG in Fig.~\ref{fig:balance_benefit}(a), we find that the change of QG is small from $w=0$ to $0.25$ while the shift in CR is small from $w=0.25$ to $1$. Thus, the weight that achieves holistic maximization
is around 0.25.
This analysis also holds for kCR / kQG and nDCG-CR / nDCG-QG.

\subsubsection{Efficiency}
We show the updating time of each method in Table~\ref{table:efficiency}. Random and Greedy CS are not included because they do not have a model to update. Taskrec and Greedy NN are supervised learning-based methods which update the whole model with incremental data. During the entire process, although we train them with newly collected data once at the end of each day, the average updating time during the whole process is still longer than 3s. LinUCB and DDQN are reinforcement learning-based methods, which update the existing model quickly after collecting every new feedback. The average updating time is in the order of milliseconds, which satisfies the real-time requirement. 

\begin{figure*}[t!]
\centering  
\includegraphics[width=\linewidth]{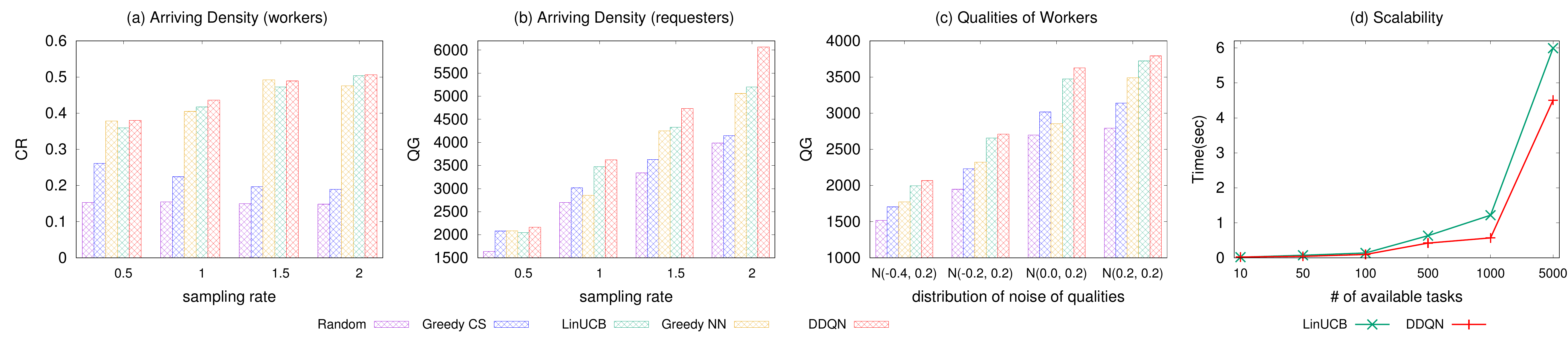}
\caption{Synthetic Results}
\label{fig:synthetic_result} 
\end{figure*}
\subsection{Experimental Results (synthetic dataset)}

\subsubsection{Arriving density of workers}
We change the number of worker arrivals ($50$k) in the real dataset using sampling with replacement. 
We range the sampling rate of worker arrivals from $0.5$ to $2.0$, resulting in $25$k to $100$k arrivals. 
For the same arrival which is sampled multiple times, we add a delta time following a normal distribution where the mean and std are 1 day, to make their arrival times distinct. 

Fig.~\ref{fig:synthetic_result}(a) and~\ref{fig:synthetic_result}(b) show the change of CR / QG with a different sampling rate of worker arrivals. Because CR is divided by the number of timestamps (i.e., the number of worker arrivals), the values of all the methods are similar at different sampling rates. QG is the absolute value, so the values of all the methods increase at a high sampling rate. The performance of our algorithm DDQN is typically better than that of others for both CR and QG in the different cases. 


\subsubsection{Distribution of qualities of workers}
We change the qualities of workers in the real dataset by adding  noise. We generate the noise from a normal distribution and add it to the original quality of workers randomly. We tried four normal distributions: $\mathcal{N}(-0.4,0.2)$, $\mathcal{N}(-0.2,0.2)$, $\mathcal{N}(0.0,0.2)$ and $\mathcal{N}(0.2,0.2)$.
The result is shown in Fig.~\ref{fig:synthetic_result}(c). Since the quality of workers only affects the quality gain of tasks, we show the change of QG for various worker qualities. Obviously, the sum of qualities of tasks becomes larger as the quality of workers increases. Moreover, DDQN always performs better than its competitors, no matter whether the worker qualities are low or high.  

\subsubsection{Scalability}

The update cost is mainly determined by the number of available tasks in RL-based methods (LinUCB and DDQN). We vary the number of the currently available tasks from $10$ to $5$k and measure the update cost in Fig.~\ref{fig:synthetic_result}(d). The plot shows that the cost is approximately linear to the number of available tasks. DDQN always spends less time than LinUCB. 
The number of available tasks at
Amazon MTurk, which is the largest platform, 
is about $1$k.
DDQN can update in real-time (around 0.5s) using one GPU for $1$k tasks.
Parallel computation with multiple GPUs can be used to support an even higher number of tasks.







\section{related work}\label{sec:relatedwork}

\subsection{Reinforcement learning and deep reinforcement learning}

Unlike supervised learning which requires labeled training data and infers a classification or a regression model, reinforcement learning (RL) learns how agents should take sequences of actions in an unknown environment in order to maximize cumulative rewards. The environment is formulated as a Markov Decision Process \cite{bellman1957markovian}, and the agent makes a tradeoff between exploring untouched space and exploiting current knowledge.
RL methods are mainly divided into three categories, model-free, model-based and policy search, based on the assumption of MDPs. In this paper, we utilize the model-free method, Q-learning \cite{Q_learning}, which estimates a Q-function iteratively using Bellman backups \cite{sutton2018reinforcement} and acts greedily based on Q-functions until convergence.

Deep reinforcement learning is a combination of RL and deep learning. Deep RL has experienced dramatic growth recently in multiple fields, including games (AlphaGo)~\cite{epsilon_greedy, double_q_learning, dueling}, robotics~\cite{robotic}, natural language processing~\cite{nlp1, dialogue}, computer vision~\cite{cv1,cv2}, finance~\cite{financial}, computer systems~\cite{mao2016resource, tuning1, tuning2}, recommender systems~\cite{zheng2018drn, zhao2018recommendations,zou2019reinforcement} and so on. Deep Q-Network (DQN) is an improved version of Q-learning with a neural network. The applications of DQN in recommender systems \cite{zheng2018drn, zhao2018recommendations} are the most related to our paper. Instead of recommending items to users, we arrange tasks to workers. However, recommender systems only consider the benefit of users, which is just one objective of our framework.

\subsection{Task Recommendation and Assignment in Crowdsourcing} 

\subsubsection{supervised learning}
Significant research on task and worker recommendation using supervised learning has been developed during the past few years. Content-based recommendation methods \cite{ambati2011towards,yuen2012task,satzger2011stimulating} match task profiles to worker profiles. They use features of workers and tasks (e.g., a bag of words from user profiles) and the task selection history or worker's performance history. They calculate similarity and recommend based on these features. Collaborative filtering has also been used in crowdsourcing. For example, \cite{taskrec} builds the task-worker, worker-category and task-category matrices, and applies probabilistic matrix factorization to capture workers' preferences. \cite{safran2018efficient} uses category-based matrix factorization and $k$NN algorithms to recommend top-$k$ tasks to workers. 

\subsubsection{reinforcement learning}

Some studies have applied reinforcement learning for  spatial crowdsourcing \cite{spatialUCB, RLBipartiteGraphMatching}. \cite{spatialUCB} proposes a multi-armed bandit approach for online spatial task assignment. The task acceptance rate of the worker is modeled as a linear model of the travel distance and task type, and the goal is to maximize the cumulative success rate of assignments. 
In \cite{RLBipartiteGraphMatching}, an RL-based algorithm is proposed to solve a dynamic bipartite graph matching problem. However, a simple state representation is used, i.e., the number of available nodes in the bipartite graph, which limits the power of RL. 


\section{conclusions}\label{sec:conclusions}
In this work, we propose a novel Deep Reinforcement Learning framework for task arrangement in crowdsourcing platforms. We consider the benefits of workers and requesters simultaneously to help the platforms to attract more tasks and workers and achieve profit maximization. We also utilize a Deep Q-Network paired with novel and effective representations of state, action, reward, state transition and future state, and revised equations for deriving Q values. 
Experiments on both real and synthetic datasets verify the effectiveness and efficiency of our framework. 

There are two future directions to consider. First, we can apply alternative deep RL methods, such as deep deterministic policy gradient. This method can project the list of tasks into a continuous action space and obtain more accurate sorting. Another issue is how to handle conflicts when two workers come almost at the same time. It is hard to model the situation that a worker comes while previous workers are still completing tasks and have not given their feedback. Our current solution ignores any unknown completions from previous workers.
In the future, we can adapt our model and 
consider these assigned but not completed tasks, to better improve the quality of the task arrangement.    
\section*{Appendix}

\begin{definition} (Permutation-invariant Function)
Let $\{\sigma\}$ be the set of all permutations of indices $\{1,..,n\}$. A function of $f: X^{n} \to Y^{n}$ is permutation-invariant iff for any permutation in $\{\sigma\}$, $f(\sigma x)=\sigma f(x)$.
\end{definition}
\begin{proof} (\textbf{rFF function is Permutation-invariant.}) \\
Let $\boldsymbol{X}= 
\left[
\begin{matrix}
\boldsymbol{x_1} \\
 \vdots \\
\boldsymbol{x_n} \\
\end{matrix}
\right]$, where each row is the feature of an item in the set. Then, rFF($\boldsymbol{X}$) = 
$ 
\text{relu}(\boldsymbol{X} W + b )=  
\left[
\begin{matrix}
\text{relu}(\boldsymbol{x_1}  W + b ) \\
 \vdots \\
\text{relu}(\boldsymbol{x_n}  W + b ) \\
\end{matrix}
\right]
$. 
The value in row $i$ of rFF($\boldsymbol{X}$) only depends on $\boldsymbol{x_i}$ and is independent to $\boldsymbol{x_j} \text{ where } \forall j \neq i$.
\end{proof}

\begin{proof} (\textbf{MultiHead Self-Attention Layer is Permutation -invariant.}) \\
First of all, we prove that each $\text{head}_j= \text{Att}(X W^{Q}_j , X W^{K}_j, X W^{V}_j)$ is permutation-invariant. \\
Similarly, let $\boldsymbol{X}= 
\left[
\begin{matrix}
\boldsymbol{x_1} \\
 \vdots \\
\boldsymbol{x_n} \\
\end{matrix}
\right]
$ and $W^{Q}_{j}(W^{K}_{j})^{T} = W'_{j},$ then
$
X W^{Q}_{j} (X W^{K}_{j})^{T} = X W'_{j} X^T = 
\left[
\begin{matrix}
\boldsymbol{x_1} W'_{j} \boldsymbol{x_1^T},  \cdots,  \boldsymbol{x_1} W'_{j} \boldsymbol{x_n^T} \\
 \vdots \\
\boldsymbol{x_n} W'_{j} \boldsymbol{x_1^T},  \cdots,  \boldsymbol{x_n} W'_{j} \boldsymbol{x_n^T} \\
\end{matrix}
\right]
$
. After multiplying $X W^{V}_j$ and scaling by 
$\omega(\cdot)$,
$\text{head}_j$ becomes
$
\left[
\begin{matrix}
\sum_{i=1}^{n} \omega(\boldsymbol{x_1} W'_{j} \boldsymbol{x_i^T})  \boldsymbol{x_i} W^{V}_j\\
 \vdots \\
\sum_{i=1}^{n} \omega(\boldsymbol{x_n} W'_{j} \boldsymbol{x_i^T})  \boldsymbol{x_i} W^{V}_j \\
\end{matrix}
\right]
$.
Each value in row $i$ of $\text{head}_j$ depends on $\boldsymbol{x_i}$ and weighed sum of $\boldsymbol{x_j},  \forall j$, which is also permutation-invariant. 

Next we consider  $\text{MultiHead}(X , X, X)$. Because of $\text{Concat}(\text{head}_{1}, ...,\text{head}_{h})$ and multiplying $W^{O}$ are both row-wise, we can prove the permutation-invariance in the same way as for the  rFF function. 
\end{proof}

{ 
\small
\bibliographystyle{abbrv} 
\bibliography{paper}
}

\end{document}